%% file: main.tex
\documentclass[10pt,a4paper]{article}

\usepackage{technicalreport}
\usepackage{pifont}
\usepackage{xspace}
\usepackage{bm}

\newcommand{\bench}{\textsc{MarrBench}\xspace}
\DeclareRobustCommand{\famid}[1]{\textrm{\textsc{#1}}}
\DeclareRobustCommand{\taskid}[1]{\textrm{\textsc{#1}}}

\newcommand{\cmark}{\ding{51}}
\newcommand{\xmark}{\ding{55}}
\newcommand{\pmark}{$\triangle$}
\definecolor{bestred}{HTML}{C01A1A}
\definecolor{secondblue}{HTML}{1B5E9E}
\newcommand{\best}[1]{\textbf{\textcolor{bestred}{#1}}}
\newcommand{\second}[1]{\textbf{\textcolor{secondblue}{#1}}}

\DeclareMathOperator{\Exec}{Exec}
\DeclareMathOperator{\clip}{clip}
\newcommand{\Mm}{\mathcal{M}}
\newcommand{\Vv}{\mathcal{V}}
\newcommand{\Ll}{\mathcal{L}}
\newcommand{\Tt}{\mathcal{T}}
\newcommand{\Yhat}{\hat{Y}}
\newcommand{\yhat}{\hat{y}}

\makeatletter
\DeclareRobustCommand\onedot{\futurelet\@let@token\@onedot}
\def\@onedot{\ifx\@let@token.\else.\null\fi\xspace}
\def\eg{\emph{e.g}\onedot}

 \def\vs{\emph{vs}\onedot}
 
\makeatother

\newcommand{\instlogo}[2]{\ensuremath{\vcenter{\hbox{\includegraphics[height=#1]{#2}}}}}
\newcommand{\ghlogo}{\raisebox{-0.16em}{\includegraphics[height=1.0em]{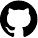}}}
\newcommand{\hflogo}{\raisebox{-0.20em}{\includegraphics[height=1.05em]{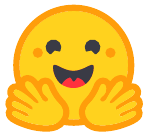}}}

\headerlogos{%
  \instlogo{8.4mm}{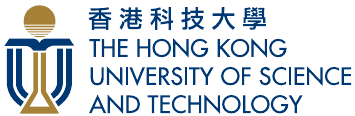}\hfill
  \instlogo{8.4mm}{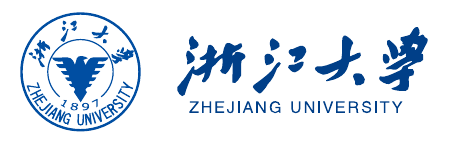}\hfill
  \instlogo{6.5mm}{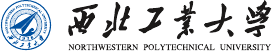}\hfill
  \instlogo{8.2mm}{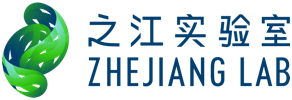}\hfill
  \instlogo{8.4mm}{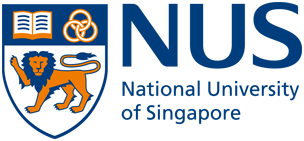}\hfill
  \instlogo{7.8mm}{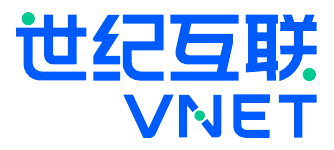}%
}

\title{\texorpdfstring{MarrBench:\\Can Multimodal Agents Solve Classic Vision Tasks?}{MarrBench: Can Multimodal Agents Solve Classic Vision Tasks?}}

\author{%
  Jiaqi Tang\textsuperscript{1,*}\quad
  Xiaogang Xu\textsuperscript{2,*,\ensuremath{\dagger}}\quad
  Youyang Zhai\textsuperscript{3,*}\quad
  Yaoyu Yang\textsuperscript{3,*}\quad
  Jianmin Chen\textsuperscript{3,*}\quad
  Botong Geng\textsuperscript{3,*}\\
  Yingying Yan\textsuperscript{3,*}\quad
  Lihao Liu\textsuperscript{3,*}\quad
  Wei Wei\textsuperscript{3}\quad
  Kelu Yao\textsuperscript{4}\quad
  Xiaobin Hu\textsuperscript{5}\\
  Aimin Pan\textsuperscript{6}\quad
  Jiafei Wu\textsuperscript{2}\quad
  Zhe Liu\textsuperscript{2}\quad
  Qifeng Chen\textsuperscript{1,\ensuremath{\ddagger}}\quad
  Shuicheng Yan\textsuperscript{5}%
}

\affiliation{%
  \textsuperscript{1}The Hong Kong University of Science and Technology\quad
  \textsuperscript{2}Zhejiang University\quad
  \textsuperscript{3}Northwestern Polytechnical University\\
  \textsuperscript{4}Zhejiang Lab\quad
  \textsuperscript{5}National University of Singapore\quad
  \textsuperscript{6}VNET Group%
}

\authornotes{%
  \textsuperscript{*}Equal contribution.\qquad
  \textsuperscript{\ensuremath{\dagger}}Project leader.\qquad
  \textsuperscript{\ensuremath{\ddagger}}Corresponding author.%
}

\reportpdfauthor{Jiaqi Tang, Xiaogang Xu, Youyang Zhai, Yaoyu Yang, Jianmin Chen, Botong Geng, Yingying Yan, Lihao Liu, Wei Wei, Kelu Yao, Xiaobin Hu, Aimin Pan, Jiafei Wu, Zhe Liu, Qifeng Chen, Shuicheng Yan}
\date{}

\begin{document}

\maketitle

\input{sections/00-abstract}

\begin{center}
  \includegraphics[width=\linewidth]{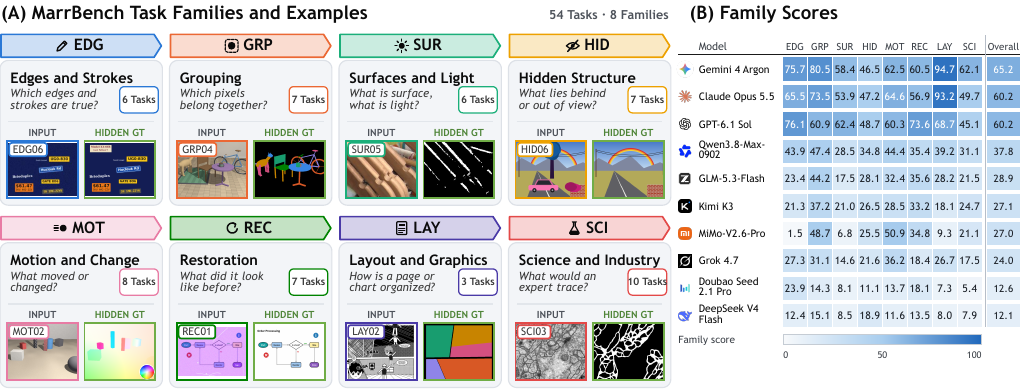}
  \captionsetup{hypcap=false}
  \captionof{figure}{\textbf{\bench evaluates multimodal agents across eight classic-vision task families.}
  (A)~One input/ground-truth pair per family (54 tasks; real: SHIQ~\cite{fu2021shiq}, ISBI-2012~\cite{arganda2012isbi}).
  (B)~Family and overall scores.}
  \captionsetup{hypcap=true}
  \label{fig:teaser}
\end{center}
\vspace{2mm}

\input{sections/01-introduction}
\input{sections/02-related-work}
\input{sections/03-benchmark}
\input{sections/04-experiments}
\input{sections/05-conclusion}

\FloatBarrier
\clearpage
{\small
\setlength{\bibsep}{2.2pt plus 0.5pt minus 0.5pt}
\bibliographystyle{unsrtnat}
\bibliography{references}
}

\end{document}

%% file: sections/00-abstract.tex
\begin{abstract}
Multimodal agents increasingly do visual work, yet their vision is evaluated by answering questions about images or by computing results with operators and pretrained models, which credits a stated answer or the tools rather than the agent's own perception.
We introduce \bench, which poses classic vision tasks to generalist agents under one rule: the image reaches only the model's vision encoder, and the result map must be drawn by a program the agent writes, re-executed without images and scored against hidden ground truth.
The suite comprises 54 tasks in eight families: detecting edges and restoring strokes; grouping pixels into parts and instances; separating surfaces, reflectance, and illumination; inferring occluded structure; estimating motion and cross-frame correspondence; restoring warped or degraded images; parsing layouts and graphics; and extracting structures from scientific imagery.
Each task is scored by two to five checkpoints with fixed task-specific weights that prioritize the defining pixel geometry of dense outputs.
Across ten high-effort multimodal models, Gemini 4 Argon leads with 65.2/100, followed by Claude Opus 5.5 and GPT-6.1 Sol at 60.2.
Every model earns 14.0--36.4 points more coarse- than pixel-level credit within the same tasks, so recognition remains substantially easier than pixel delineation.

\par\vspace{2.5mm}
\noindent\begin{tabular}{@{}l@{\hspace{0.75em}}l@{}}
  \textbf{Harness:} & \href{https://github.com/OceanMind-Lab/MarrBench}{\ghlogo\,\texttt{\footnotesize https://github.com/OceanMind-Lab/MarrBench}} \\
  \textbf{Benchmark:} & \href{https://huggingface.co/datasets/OceanMind-Lab/MarrBench}{\hflogo\,\texttt{\footnotesize https://huggingface.co/datasets/OceanMind-Lab/MarrBench}}
\end{tabular}
\end{abstract}

%% file: sections/01-introduction.tex
\section{Introduction}
\label{sec:intro}

Multimodal agents increasingly do visual work: they operate graphical interfaces~\cite{xie2024osworld,liu2025visualagentbench}, reason over images by writing code~\cite{zhao2025pyvision,google2026agenticvision}, and reproduce rendered webpages and charts as programs~\cite{si2025design2code,yang2025chartmimic}.
Whether such an agent can solve classic computer vision tasks itself, delineating the boundaries, instances, or changes in an unseen image as a careful annotator would, remains an open question.
Answering it requires a score that can be attributed to what the agent perceived rather than to the tools it called.

\begin{figure}[htbp]
  \centering
  \includegraphics[width=0.62\linewidth]{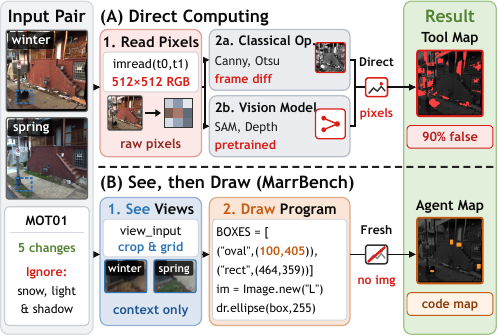}
  \caption{\textbf{Direct Computing versus See, then Draw.}
  (A)~Direct computing reads raw pixels via classical operators or specialist models (\eg, Canny~\cite{canny1986}, SAM~\cite{kirillov2023sam}), crediting the tool.
  (B)~\bench shows only rendered views and re-runs the agent's program in an image-free container.}
  \label{fig:motivation}
\end{figure}

Existing evaluations of an agent's vision primarily follow two paradigms.
\emph{Answering} reduces perception to a choice or a short reply, from classic perception problems recast as multiple-choice questions~\cite{fu2024blink,tong2024eyes,tong2024cambrian,liu2025obench} to multi-step visual questions solved with tools~\cite{zhu2026vtcbench,su2026agentvista}.
\emph{Computing} lets code crop, filter, and measure the pixels, or lets pretrained models produce the result~\cite{gupta2023visprog,suris2023vipergpt,yang2026cwi,wang2026vasa,hao2026segagent}.
While effective for their purposes, they suffer from two fundamental limitations:
\textbf{(i) Answering is not producing:} a correct answer certifies recognition, not delineation; a model may name a change correctly, or simply guess it, without being able to mark the pixels that changed.
\textbf{(ii) Confounded attribution:} once code reads pixels or calls a vision model, the result map can be computed on a \emph{direct route} that bypasses the agent's understanding, such as frame differencing for change detection~\cite{radke2005change}, an edge detector~\cite{canny1986} for boundaries, or SAM~\cite{kirillov2023sam} for instance masks, so the score grades the tools rather than the agent (\cref{fig:motivation}-A).

To overcome these limitations, we introduce \bench\footnote{Named in honor of David Marr (1945--1980) and his foundational contributions to computer vision.}, a benchmark that poses classic vision tasks to a generalist multimodal agent under a single rule: the only path from input pixels to the scored result runs through the agent's own perception (\cref{fig:motivation}-B).
Input images stay on the host and reach the model only as rendered views; the agent writes a program that \textbf{draws the full result map}; and the program is re-executed in a fresh container without images, where only its output is scored against hidden ground truth.
The direct route is thus removed by construction rather than by inspection, and the score reflects what the agent saw and could render.

Our benchmark is built around three core components (\cref{fig:overview}).
\textbf{First}, a closed-book environment (\cref{sec:env}): a network-isolated sandbox with general-purpose scientific libraries but no vision model, labels, or tests, paired with a host-side view tool that renders crops, grids, channel and contrast adjustments, and overlays of the agent's own output into the conversation context.
\textbf{Second}, a comprehensive task suite (\cref{sec:tasks,sec:data}): organized after Marr's progression from edges to surfaces to objects~\cite{marr1982vision}, the suite comprises 54 tasks in eight families, from edges and grouping to hidden structure and scientific imagery (\cref{fig:teaser,tab:suite}).
\textbf{Finally}, checkpoint scoring (\cref{sec:scoring}): each task is scored by two to five checkpoints that each measure one property, with fixed task-specific weights emphasizing the defining pixel geometry of dense outputs.

We evaluate ten high-effort multimodal models across all 54 tasks through one shared agent interface (\cref{sec:setup}).
Gemini 4 Argon~\cite{google2026gemini4argon} leads with $\bar S=65.2$, followed by Claude Opus 5.5~\cite{anthropic2026opus55} and GPT-6.1 Sol~\cite{openai2026sol} at 60.2 (\cref{tab:main}).
Across every model, coarse checkpoint credit exceeds pixel-level credit within the same tasks by 14.0--36.4 points.
Therefore, the central bottleneck is not merely recognizing the right scene content, but rendering it at pixel-accurate geometry (\cref{fig:results,eq:tier-contrast}).
Our contributions are summarized as follows:
\begin{itemize}
\item We introduce \bench, 54 classic vision tasks in eight families in which input images reach only the agent's vision encoder and the result is drawn by a program re-executed without images, removing the direct pixel route by construction.
\item We design a closed-book evaluation environment and a 242-checkpoint scoring protocol with fixed task-specific weights prioritizing pixel geometry.
\item We evaluate ten frontier multimodal models across all 54 tasks, revealing that while the best reaches 65.2/100, all models remain far stronger at coarse recognition than at pixel-accurate delineation.
\end{itemize}

%% file: sections/02-related-work.tex
\section{Related Work}
\label{sec:related}

\input{tables/comparison}

\Cref{tab:comparison} contrasts \bench with representative benchmarks along the properties that define it.

\paragraph{Perception Benchmarks for MLLMs.}
BLINK~\cite{fu2024blink}, MMVP~\cite{tong2024eyes}, and CV-Bench~\cite{tong2024cambrian} recast classic perception problems, such as correspondence, relative depth, and counting, as multiple-choice questions, and O-Bench~\cite{liu2025obench} poses such questions about occluded objects; strong models often fail them.
Agentic perception benchmarks extend this question-answering paradigm to multi-step tool use~\cite{titan2025cvagent,su2026agentvista,zhu2026vtcbench} and interactive environments~\cite{liu2025visualagentbench,xie2024osworld,cao2026agenticvbench,lin2026cvarena}.
However, these benchmarks grade a stated choice or a short reply rather than a delineated perceptual map.
In contrast, \bench grades the rendered output, a dense map in 45 of the 54 tasks, against hidden ground truth: a model that picks or states the right relation may still be unable to draw where that relation holds in the image.

\paragraph{Classic Vision Tasks.}
To obtain dense predictions on classic vision tasks, several works train multimodal models to output masks~\cite{lai2024lisa}, dense maps without task-specific decoders~\cite{li2026densemllm,han2026sensenova}, metric depth~\cite{cai2026depthlm}, or amodal masks of occluded objects~\cite{li2025aura} using human annotations or 3D data~\cite{zhu2017cocoa,qi2019kins,zhan2024amodal}.
Other approaches couple a multimodal model with a segmentation foundation model through learned clicks~\cite{zhu2025segagent} or training-free mask and box refinement~\cite{wang2026vasa,hao2026segagent}, or prompt image and video generators to produce depth, normals, albedo, edges, and segmentation zero-shot~\cite{yang2026physeditbench,liu2026editors,wiedemer2025video}.
In all these settings a specialized or generative vision network computes the map directly from the input pixels.
In contrast, \bench provides no auxiliary vision model and no task training: only the evaluated agent's vision encoder sees the rendered views, and the map must be produced from what the agent itself perceived.

\paragraph{Visual Coding.}
Visual programming composes pretrained vision modules through generated code~\cite{gupta2023visprog,suris2023vipergpt}, Visual Sketchpad lets models draw intermediate sketches~\cite{hu2024sketchpad}, and ``thinking with images'' interleaves reasoning with image cropping and filtering through native tools~\cite{openai2025thinking} or code~\cite{google2026agenticvision,zhao2025pyvision,zhang2025thyme,guo2026codevision,hou2026codev}.
In code-based benchmarks, CwI-Bench~\cite{yang2026cwi} lets Python code process the image to output a number, Blueprint-Bench~\cite{petersson2025blueprint} scores floor plans by room-graph connectivity, Design2Code~\cite{si2025design2code}, ChartMimic~\cite{yang2025chartmimic}, and VIGA~\cite{yin2026viga} translate webpages, charts, or scenes into code, and coding or ML benchmarks verify agents with executable tests~\cite{jimenez2024swebench,merrill2026terminalbench}, terminal media tools~\cite{heo2026mmtb}, or labeled training splits~\cite{chan2025mlebench}.
Whenever code can read the input pixels or call a vision model, the score confounds perception with the tool's pixel computation.
In contrast, \bench asks the agent to write code that draws a classic vision output, often hidden structure such as a wall behind furniture, while keeping code completely blind to the input pixels and providing neither tests nor labels.

%% file: tables/comparison.tex
\begin{table}[htbp]
\centering
\footnotesize
\setlength{\tabcolsep}{3.5pt}
\renewcommand{\arraystretch}{1.12}
\caption{\textbf{Comparison with representative benchmarks.}
\cmark/\pmark/\xmark: yes/partially/no; --: not applicable (no code written). Traj.\ replay: intermediate states are re-scored against the hidden ground truth.
In \bench, 45 of the 54 tasks evaluate dense per-pixel maps.}
\label{tab:comparison}
\begin{tabularx}{\linewidth}{@{}l >{\raggedright\arraybackslash}X *{7}{c}@{}}
\toprule
\multirow{2.3}{*}{\textbf{Category}} & \multirow{2.3}{*}{\textbf{Benchmark}}
  & \multicolumn{2}{c}{\textbf{Task and output}}
  & \multicolumn{3}{c}{\textbf{Closed-book rule}}
  & \multicolumn{2}{c}{\textbf{Measurement}} \\
\cmidrule(lr){3-4} \cmidrule(lr){5-7} \cmidrule(l){8-9}
 & & Classic & Dense & No aux. & Code & No task & Pixel & Traj. \\[-0.25ex]
 & & tasks & output & model & blind & training & scoring & replay \\
\midrule
\multirow{3}{*}{Perception benchmarks}
 & BLINK~\cite{fu2024blink} & \cmark & \xmark & \cmark & --     & \cmark & \xmark & \xmark \\
 & MMVP~\cite{tong2024eyes}, CV-Bench~\cite{tong2024cambrian}
 & \cmark & \xmark & \cmark & --     & \cmark & \xmark & \xmark \\
 & O-Bench~\cite{liu2025obench} & \pmark & \xmark & \cmark & --     & \cmark & \xmark & \xmark \\
\midrule
\multirow{4}{*}{Classic vision tasks}
 & PhysEditBench~\cite{yang2026physeditbench} & \cmark & \cmark & \cmark & --     & \pmark & \cmark & \xmark \\
 & Blueprint-Bench~\cite{petersson2025blueprint} & \pmark & \pmark & \cmark & \pmark & \cmark & \xmark & \xmark \\
 & CV-Arena~\cite{lin2026cvarena} & \pmark & \pmark & \pmark & --     & \cmark & \xmark & \xmark \\
 & MLE-bench~\cite{chan2025mlebench} & \pmark & \pmark & \xmark & \xmark & \xmark & \pmark & \pmark \\
\midrule
\multirow{5}{*}{Visual coding}
 & Titan CV Agent~\cite{titan2025cvagent} & \pmark & \xmark & \xmark & \xmark & \cmark & \xmark & \xmark \\
 & VTC-Bench~\cite{zhu2026vtcbench} & \pmark & \xmark & \cmark & \xmark & \cmark & \xmark & \xmark \\
 & CwI-Bench~\cite{yang2026cwi} & \pmark & \xmark & \cmark & \xmark & \pmark & \xmark & \xmark \\
 & AgenticVBench~\cite{cao2026agenticvbench} & \pmark & \pmark & \pmark & \xmark & \cmark & \pmark & \xmark \\
 & Design2Code~\cite{si2025design2code}, ChartMimic~\cite{yang2025chartmimic}
 & \xmark & \pmark & \cmark & \cmark & \cmark & \xmark & \xmark \\
\midrule
\rowcolor{reportLightBlue}
\textbf{Ours} & \textbf{\bench} & \cmark & \cmark & \cmark & \cmark & \cmark & \cmark & \cmark \\
\bottomrule
\end{tabularx}
\end{table}

%% file: sections/03-benchmark.tex
\section{\texorpdfstring{The \bench Benchmark}{The MarrBench Benchmark}}
\label{sec:benchmark}

\begin{figure}[htbp]
  \centering
  \includegraphics[width=\linewidth]{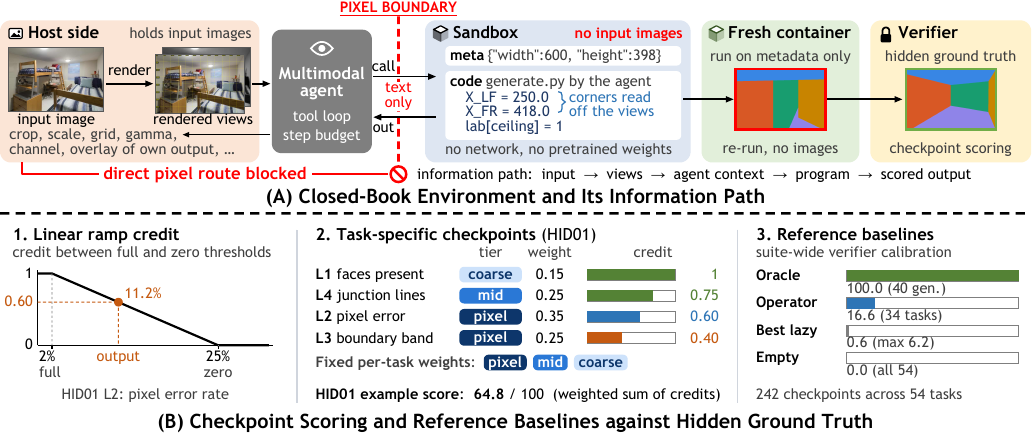}
  \caption{\textbf{Overview of \bench.}
  \textbf{(A)}~Closed-book environment (LSUN Room Layout~\cite{yu2015lsun}): images stay on the host as rendered views; only text enters the sandbox; the program is re-executed without images.
  \textbf{(B)}~Checkpoint scoring: fixed pixel-dominant weighted ramps, calibrated against oracle, operator, lazy, and empty baselines (weights and credits of the \taskid{HID01} example are illustrative).}
  \label{fig:overview}
\end{figure}

\subsection{Formulation and Attribution}
\label{sec:problem}

A classic computer vision task $\tau$ defines a mapping $f_\tau: \mathcal{X}_\tau \to \mathcal{Y}_\tau$ from an input visual domain $\mathcal{X}_\tau$ (single images, stereo/temporal pairs, video sequences, or posed multi-view sets) to a 2D structured target space $\mathcal{Y}_\tau$, comprising dense per-pixel maps in $\mathbb{R}^{H\times W\times C}$ (semantic/instance labels, binary masks, alpha mattes, optical flow, or intrinsic fields) or parametric geometric structures (wireframes, poses, homographies, unwarping grids, or frame orderings).
In \bench, the suite $\Tt = \bigcup_{F \in \mathcal{F}} \Tt_F$ ($|\Tt|=54$ tasks across $|\mathcal{F}|=8$ families) instantiates each task as a four-tuple:
\begin{equation}
\begin{aligned}
\tau &= (\iota, X, m, Y) \in \Tt, \qquad f_\tau: \mathcal{X}_\tau \to \mathcal{Y}_\tau,\\
X &= \{x_i\}_{i=1}^{n_\tau} \subset \mathcal{X}_\tau, \qquad\;\; Y = \{y_i = f_\tau(x_i)\}_{i=1}^{n_\tau} \subset \mathcal{Y}_\tau,
\end{aligned}
\label{eq:task}
\end{equation}
where $\iota$ is the task specification (problem statement, output schema, and evaluation criteria), $X$ is the set of $n_\tau$ test inputs, $m = \{m_i\}_{i=1}^{n_\tau}$ is the non-image metadata (image dimensions, frame IDs, camera parameters), and $Y = \{y_i\}_{i=1}^{n_\tau}$ is the hidden ground truth withheld on the host.
$f_\tau$ is the idealized task mapping: for occluded or procedurally generated targets, $y_i$ is defined by the generating scene (\cref{eq:data-pipeline}) rather than recoverable from $x_i$ alone.

\paragraph{Confounded Attribution.}
In conventional tool- or code-augmented evaluations, an agent policy $\pi_{\Mm}$ operates in an environment where generated code can directly read $X$ or invoke external operators and pretrained vision models $g \in \mathcal{G}$ (\eg, Canny~\cite{canny1986}, Otsu~\cite{otsu1979}, an optical-flow solver~\cite{farneback2003}, or SAM~\cite{kirillov2023sam}) to produce $\Yhat^{\mathrm{dir}} = g(X; a_{1:T})$:
\begin{equation}
\mathrm{I}\big(X;\, \Yhat^{\mathrm{dir}} \mid \iota, m\big)
\le \mathrm{I}\big(X;\, \mathbf{V}_T \mid \iota, m\big)
+ \mathrm{I}\big(X;\, g(X; a_{1:T}) \mid \mathbf{V}_T, \iota, m\big),
\label{eq:direct}
\end{equation}
where $\mathbf{V}_T$ denotes the visual views inspected by $\Mm$.
Whenever the second term in \cref{eq:direct} is positive, the score conflates the perception of $\Mm$ with the pixel computation of $g$ (\cref{fig:motivation}-A).
This identifies the intervention: keep $X$ on the host and expose it to the sandbox only through the agent's visual state.

\paragraph{Decoupled Formulation.}
To eliminate tool leakage by construction, \bench decouples host-side visual observation from an image-free sandbox (\cref{fig:overview}-A).
Starting from $h_0 = (\iota, m)$, the agent $\Mm$ evolves state $h_t = (h_{t-1}, a_t, o_t)$ over $t = 1,\dots,T$ steps ($T \le T_{\max}$):
\begin{equation}
\begin{aligned}
a_t &\sim \pi_{\Mm}(\cdot \mid h_{t-1}), \qquad h_t = (h_{t-1}, a_t, o_t),\\
o_t &=
\begin{cases}
\Vv(x_i; \theta_t) \in \mathcal{I}_{\mathrm{ctx}}, & \text{if } a_t \in \mathcal{A}_{\mathrm{view}},\\
\operatorname{Sbx}(a_t; m) \in \Sigma^\ast, & \text{if } a_t \in \mathcal{A}_{\mathrm{sbx}},
\end{cases}
\end{aligned}
\label{eq:interact}
\end{equation}
where view actions $a_t \in \mathcal{A}_{\mathrm{view}}$ query the host renderer $\Vv(x_i; \theta_t)$ (\cref{sec:env}) to inject a rendered image solely into the visual context $\mathcal{I}_{\mathrm{ctx}}$, while sandbox actions $a_t \in \mathcal{A}_{\mathrm{sbx}}$ execute inside a container holding only $m$ and return text in $\Sigma^\ast$.

\paragraph{Image-Free Re-Execution.}
At termination, only the final program $P \in \Sigma^{\le B}$ is retained and re-executed from scratch in a fresh container mounted with $m$ alone:
\begin{equation}
\label{eq:exec}
\begin{aligned}
\Yhat &= \Exec(P, m) = \{\yhat_i\}_{i=1}^{n_\tau}, \qquad P \in \Sigma^{\le B},\\
\mathrm{I}(X;\, \Yhat \mid \iota, m)
&\le \mathrm{I}(X;\, \mathbf{V}_T \mid \iota, m),\\
\mathrm{I}(X;\, \Yhat \mid \iota, m)
&\le \mathrm{H}(P \mid \iota, m) \le B\log_2|\Sigma|.
\end{aligned}
\end{equation}
Because $(X \mid \iota, m) \to \mathbf{V}_T \to h_T \to P \to \Yhat$ forms a Markov chain, the data-processing inequality~\cite{cover2006elements} makes \cref{eq:exec} an attribution rather than accuracy bound: image-dependent information must pass through the visual context and $P$, but the bound ensures neither sufficient views nor correct interpretation.
A static program check keeps $B\log_2|\Sigma| \ll \mathrm{H}(Y)$ by rejecting embedded images and dense pixel payloads; it does not certify a correct scene model, but requires detailed outputs to encode compact geometric decisions.

\subsection{Closed-Book Execution Environment}
\label{sec:env}

Three rules implement \cref{eq:interact,eq:exec} (\cref{fig:overview}-A):
\textbf{(R1) Host-resident pixels}, purging all raster and array inputs from the sandbox workspace;
\textbf{(R2) View-only perception}, returning rendered views solely to the conversation context with no test cases or labels in the workspace; and
\textbf{(R3) Stateless image-free re-execution}, running $P$ once in a clean, network-isolated container without images, where missing, malformed, or rule-violating outputs score zero.
Failures never subtract points: a run that ends before \texttt{submit} has its last program re-executed, inputs left without output files score zero, and a static check scores every input zero when the program embeds images, encoded pixel dumps, or oversized literal data.

\paragraph{Adaptive Viewing.}
One agent interface, shared by all models, lets every model inspect host-rendered whole images or crops with optional grids, photometric adjustments, and overlays of its own output, while retaining the eight most recent views in context (older views are evicted in batches of four and can be regenerated by repeating the call).
All models receive the same six tools (\texttt{bash}, \texttt{write\_file}, \texttt{read\_file}, \texttt{view\_input}, \texttt{view\_image}, \texttt{submit}); a step is one model reply, each \texttt{bash} call is capped at 120\,s, and a run ends on \texttt{submit}, the step or time limit, or five consecutive replies without a tool call.
The view operator composes overlay, crop, channel selection, photometry, resizing, grid, rotation, and flip; the long side is capped at 1536\,px, so fine detail requires a crop, and grid lines are labeled in original-image coordinates, so a coordinate read off any zoomed view enters the program unchanged.
Budgets are 60 steps/60\,min (42 tasks), 100 steps/90\,min (7 multi-frame/multi-view tasks), and 40 steps/60--90\,min (\taskid{SCI01}--\taskid{SCI05}).
The sandbox provides general-purpose scientific libraries but no input pixels, network, labels, tests, pretrained weights, or OCR.
This asymmetry preserves adaptive visual search and closed-loop correction while preventing the executable program from recovering a direct computational path to the source pixels.

\subsection{Task Taxonomy and Data Curation Pipeline}
\label{sec:tasks}
\label{sec:data}

\input{tables/suite}

\bench organizes its 54 tasks into eight families along Marr's hierarchy~\cite{marr1982vision}, from 2D primal-sketch primitives to $2.5$D surfaces, 3D occluded geometry, temporal correspondence, inverse restoration, and domain-specific structural parsing (\cref{fig:teaser,tab:suite}).
\famid{EDG} separates structural contours and strokes from incidental gradients and texture;
\famid{GRP} binds pixels into regions, instances, parts, and materials under ambiguous local cues;
\famid{SUR} disentangles reflectance and planar geometry from shadows, highlights, and mixed light;
\famid{HID} infers occluded geometry, amodal extent, and allocentric layouts without local pixel support;
\famid{MOT} tracks dense or object-level correspondence across frames and modalities;
\famid{REC} inverts warps, permutations, degradations, and compositing;
\famid{LAY} parses the regions and reading order of documents, comics, and charts;
and \famid{SCI} tests whether generalist perception transfers to microscopy, remote sensing, and industrial inspection without task training.

\paragraph{Dual-Track Curation.}
To pair real visual complexity with exact ground truth for occluded and physical structure, \bench combines a real-image track ($\Tt_{\mathrm{real}}$, 14 tasks) and a procedural track ($\Tt_{\mathrm{proc}}$, 40 tasks) with $\phi_i \sim p_\tau(\phi \mid \mathcal{C}_\tau)$:
\begin{equation}
\begin{aligned}
\tau \in \Tt_{\mathrm{real}}:\quad & x_i = \psi_\tau(\tilde{x}_i), \qquad\;\; y_i = \eta_\tau(\tilde{y}_i),\\
\tau \in \Tt_{\mathrm{proc}}:\quad & x_i = \nu_i\big(\mathcal{R}_\tau(\phi_i)\big), \qquad y_i = \Gamma_\tau(\phi_i) = \Exec(P^\star_\tau, m_i).
\end{aligned}
\label{eq:data-pipeline}
\end{equation}
The \emph{real track} adapts nine vision benchmarks~\cite{huang2018wireframe,yu2015lsun,xiang2018posecnn,hodan2018bop,cruz2021zind,pfitzmann2022doclaynet,alcantarilla2016streetview,maddalena2017sbmnet,fu2021shiq,zou2022visa} and five scientific benchmarks~\cite{ljosa2012bbbc,sirinukunwattana2017glas,arganda2012isbi,staal2004drive,ulman2017objective} via $(\psi_\tau, \eta_\tau)$ with anonymized filenames.
The \emph{procedural track} samples hidden scene parameters $\phi_i$ under acceptance constraints $\mathcal{C}_\tau$ that enforce phenomenon visibility and reject scenes solvable by trivial priors or tuned operators, pairing each task with an image-free oracle program $P^\star_\tau$.
$\Gamma_\tau$ never reads the rendered input, so $y_i$ is exact and includes what $x_i$ hides (occluded extents, backgrounds behind objects, future frames).
Acceptance requires the target phenomenon to be present and visible (\eg at least four stroke crossings in \taskid{EDG04}, at least 12\% of the free floor hidden in \taskid{HID03}); 13 generators additionally reject scenes on which a tuned classical operator or heuristic exceeds a cap of 0.12--0.45, and five cap content-free outputs at 0.10--0.12.
Generator difficulty was tuned on development runs of Kimi K3 toward 25--65 points; real samples were selected before model evaluation without using any model score.
In both tracks, file names become random hexadecimal identifiers and the metadata omit answer-bearing fields (\eg \taskid{SCI10}'s defect type or \taskid{MOT05}'s background).

\subsection{Multi-Granular Checkpoint Scoring}
\label{sec:scoring}
\label{sec:measure}

Each task $\tau$ is evaluated by $K_\tau \in \{2,\dots,5\}$ checkpoints (242 in total; \cref{fig:overview}-B).
Checkpoint $k$ maps a predicted output $\yhat_i$ and hidden target $y_i$ with error $e_{\tau,k} = e_k(\yhat_i, y_i)$ to a normalized credit $c_{\tau,k}(\yhat_i, y_i) \in [0, 1]$ via a gated linear ramp kernel:
\begin{equation}
\begin{aligned}
c_{\tau,k}(\yhat_i, y_i) &= g_{\tau,k}(\yhat_i, y_i)\cdot \rho\big(e_{\tau,k};\; e_{\tau,k}^{\mathrm{full}},\, e_{\tau,k}^{\mathrm{zero}}\big),\\
\rho(e;\, e^{\mathrm{full}}, e^{\mathrm{zero}}) &= \clip\!\left(\frac{e^{\mathrm{zero}} - e}{e^{\mathrm{zero}} - e^{\mathrm{full}}},\; 0,\; 1\right),
\end{aligned}
\label{eq:ramp}
\end{equation}
where $g_{\tau,k} \in [0, 1]$ is a multiplicative gate blocking one-sided strategies (\eg, gating recall by precision in \taskid{EDG01}).
Checkpoints follow four design rules: orthogonal error attribution after optimal Hungarian matching~\cite{kuhn1955}, trivial-prior normalization, explicit false-alarm penalties, and multiplicative gating.

\paragraph{Aggregation and Calibration.}
\label{sec:qc}
Each task specifies a fixed weight vector $\mathbf{w}_\tau \in \Delta^{K_\tau - 1}$ ($\sum_k w_{\tau,k} = 1$) that gives the largest share to strict pixel geometry on dense tasks ($70\%$ of task weight on average over the suite, against $5\%$ for coarse checkpoints).
The final vectors were fixed in a task-level audit after the main runs and are applied identically to the archived credits of every run; the as-run weights raise every model's $\bar S$ by 1.8--6.0 points and leave the ranking of all ten models unchanged (details in the supplement).
The audit leaves every threshold unchanged, so all oracles still score 100, and raises the mean pixel-weight share from 49\% to 70\%; because agents saw the as-run weights, the final scores grade the same outputs under a stricter pixel standard rather than behavior optimized for it.
With re-executed output $\yhat_i = \Exec(P_i, m_i)$, the task score $S_\tau(\Mm)$, family score $\bar{S}_F(\Mm)$, and suite score $\bar{S}(\Mm)$ on the 0--100 scale are
\begin{equation}
\begin{aligned}
S_\tau(\Mm) &= \frac{100}{n_\tau}\sum_{i=1}^{n_\tau}\sum_{k=1}^{K_\tau} w_{\tau,k}\, c_{\tau,k}(\yhat_i, y_i),\\
\bar{S}_F(\Mm) &= \frac{1}{|\Tt_F|}\sum_{\tau \in \Tt_F} S_\tau(\Mm), \quad
\bar{S}(\Mm) = \frac{1}{|\Tt|}\sum_{\tau \in \Tt} S_\tau(\Mm).
\end{aligned}
\label{eq:score}
\end{equation}
Partitioning all 242 checkpoints into coarse ($\mathrm{crs}$, 21), mid-level ($\mathrm{mid}$, 103), and pixel-level ($\mathrm{pix}$, 118) tiers over shared task subsets $\Tt_{t,\mathrm{pix}} = \Tt_t \cap \Tt_{\mathrm{pix}}$ yields the unweighted tier credit $\bar{c}_{\tau,t}(\Mm)$ and within-task paired contrast $\Delta_{t,\mathrm{pix}}(\Mm)$:
\begin{equation}
\begin{aligned}
\bar{c}_{\tau,t}(\Mm) &= \frac{1}{n_\tau |\mathcal{K}_{\tau,t}|}\sum_{i=1}^{n_\tau}\sum_{k \in \mathcal{K}_{\tau,t}} c_{\tau,k}(\yhat_i, y_i),\\
\Delta_{t,\mathrm{pix}}(\Mm) &= \frac{100}{|\Tt_{t,\mathrm{pix}}|}\sum_{\tau \in \Tt_{t,\mathrm{pix}}}\!\big(\bar{c}_{\tau,t} - \bar{c}_{\tau,\mathrm{pix}}\big).
\end{aligned}
\label{eq:tier-contrast}
\end{equation}
Suite-wide calibration across all 54 verifiers (\cref{fig:overview}-B) confirms: (P1~\emph{anchoring}) all 40 oracle programs score $100.0$ and empty outputs score $0.0$; (P2~\emph{lazy resistance}) content-free priors $\Ll_\tau$ average $0.6/100$ (at most $6.2$) and classical pixel operators average $16.6/100$ across 34 tasks; and (P3~\emph{selectivity}) 66 controlled perturbations place $78.6\%$ of lost points on their declared target checkpoints.
These checks exposed 12 verifier, metadata, and instruction defects that were fixed before the main runs (\eg \taskid{MOT04}'s metadata listed frames in a time-correlated order worth 28.9 points); after the fixes, no content-free output exceeds 15 points on any task, even when chosen with the ground truth.

%% file: tables/suite.tex
\begin{table}[htbp]
\centering
\small
\setlength{\tabcolsep}{6pt}
\renewcommand{\arraystretch}{1.15}
\caption{\textbf{Task families and representative tasks of \bench}, ordered along Marr's progression~\cite{marr1982vision}.}
\label{tab:suite}
\begin{tabularx}{\linewidth}{@{}l >{\raggedright\arraybackslash}X r@{}}
\toprule
\textbf{Family} & \textbf{Representative Tasks} & \textbf{Tasks} \\
\midrule
\textbf{\famid{EDG}} \emph{Edges \& strokes} & object boundaries, ink binarization, icon vectorization, stroke order, wireframes, text restoration & 6 \\
\textbf{\famid{GRP}} \emph{Grouping} & semantic rooms, instances, alpha matting, parts, camouflage, glass, mirrors & 7 \\
\textbf{\famid{SUR}} \emph{Surfaces \& light} & intrinsic albedo, cast/relit shadows, coplanar facets, highlights, mixed illuminants & 6 \\
\textbf{\famid{HID}} \emph{Hidden structure} & room layout, amodal completion, floor maps, novel views, occlusion graphs, inpainting & 7 \\
\textbf{\famid{MOT}} \emph{Motion \& change} & change detection, optical flow, motion prediction, frame order, tracking, SAR registration & 8 \\
\textbf{\famid{REC}} \emph{Restoration} & compound restoration, puzzle reassembly, orientation, stitching, dewarping, splicing & 7 \\
\textbf{\famid{LAY}} \emph{Layout \& graphics} & document layout, comic panels \& reading order, multi-series chart splitting & 3 \\
\textbf{\famid{SCI}} \emph{Science \& industry} & cell nuclei, glands, EM membranes, vessels, lineages, aerial land cover, cracks, PCB defects & 10 \\
\midrule
\rowcolor{reportLightBlue}
\textbf{Total} & eight task families & \textbf{54} \\
\bottomrule
\end{tabularx}
\end{table}

%% file: sections/04-experiments.tex
\section{Experiments}
\label{sec:experiments}

\subsection{Experimental Setup}
\label{sec:setup}

\input{tables/mainresults}

\paragraph{Models.}
We evaluate ten multimodal models across all 54 tasks (\cref{tab:main}): Gemini 4 Argon~\cite{google2026gemini4argon}, Claude Opus 5.5~\cite{anthropic2026opus55}, GPT-6.1 Sol~\cite{openai2026sol}, Qwen3.8-Max-0902~\cite{qwen2026max0902}, GLM-5.3-Flash~\cite{zai2026glm53flash}, Kimi K3~\cite{kimi2026k3}, MiMo-V2.6-Pro~\cite{mimo2026v26pro}, Grok 4.7~\cite{xai2026grok47}, Doubao Seed 2.1 Pro~\cite{bytedance2026seed21}, and DeepSeek V4 Flash~\cite{deepseek2026v4}.
All reported leaderboard runs use High reasoning or the provider-equivalent maximum: high for eight endpoints, Claude's adaptive mode capped at high, and reasoning on for MiMo, whose endpoint exposes only on/off (\cref{tab:main}).

\paragraph{Closed-Book Protocol.}
Every model uses the same closed-book environment (\cref{sec:env}) with one English system prompt shared by all tasks, and no model receives demonstrations or fine-tuning (details in the supplement).
A run that times out is retried once under the same budget.

\paragraph{Scoring and Statistics.}
Only the image-free re-execution of the final program is scored (\cref{eq:exec}); in-session outputs never are.
We report per-task scores $S_\tau$, family means $\bar S_F$, and the equal-weight suite mean $\bar S$ (\cref{eq:score}), with 95\% percentile-bootstrap confidence intervals over tasks~\cite{efron1994bootstrap} ($B=10{,}000$ resamples, fixed seed); model differences are resampled on the same tasks, so shared task difficulty cancels.
Each run is a long-horizon episode (26--63 mean steps per model); to keep the suite broad rather than deep, 49 tasks hold one test input (87 inputs in total; details in the supplement) and each model runs each task once, so the intervals reflect the sampling of tasks, not run-to-run variance.

\subsection{Main Results}
\label{sec:main}

\begin{figure}[htbp]
  \centering
  \includegraphics[width=\linewidth]{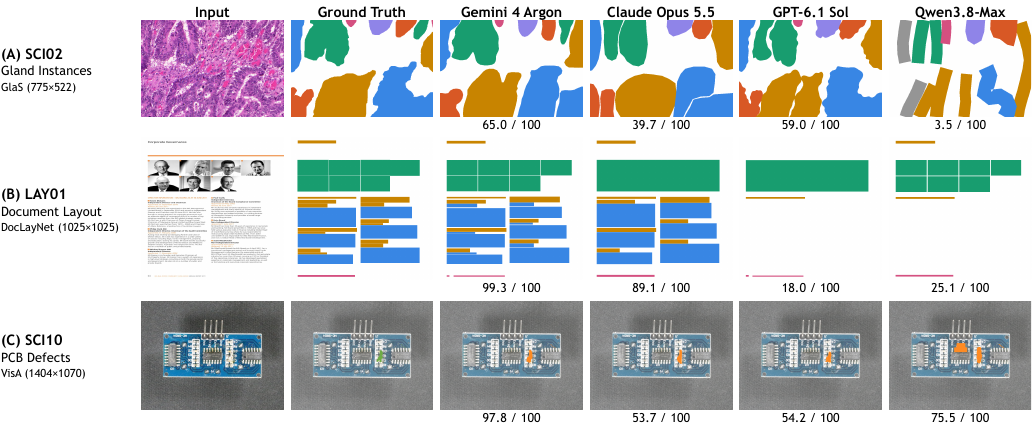}
  \caption{\textbf{Qualitative comparison of image-free re-executed outputs across frontier models} (native aspect ratios; task scores out of 100).
  \textbf{(A)}~\taskid{SCI02}/GlaS~\cite{sirinukunwattana2017glas}: Gemini/GPT-6.1/Claude trace all 11 glands; Qwen3.8 draws straight strips.
  \textbf{(B)}~\taskid{LAY01}/DocLayNet~\cite{pfitzmann2022doclaynet}: Gemini/Claude delineate 28 blocks; GPT-6.1/Qwen3.8 mislabel headers.
  \textbf{(C)}~\taskid{SCI10}/VisA~\cite{zou2022visa}: Gemini/Qwen3.8 find both defects; Claude/GPT-6.1 miss the thin trace.}
  \label{fig:qualitative}
\end{figure}

\begin{figure}[htbp]
  \centering
  \includegraphics[width=0.68\linewidth]{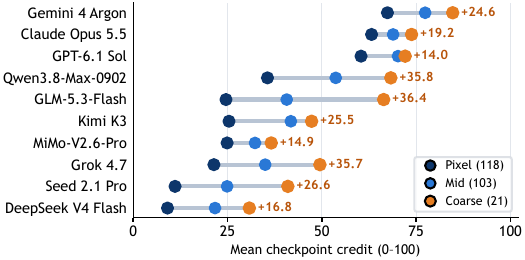}
  \caption{\textbf{Recognition versus pixel delineation across tiers.}
  Mean checkpoint credit on pixel (118), mid (103), and coarse (21) tiers; orange labels show the within-task coarse-minus-pixel gap.}
  \label{fig:results}
\end{figure}

\Cref{tab:main,fig:teaser} report family and overall scores for all ten models; \cref{fig:qualitative,fig:results} show those numbers in the rendered maps and across checkpoint tiers.

\paragraph{Frontier Cluster.}
Gemini 4 Argon~\cite{google2026gemini4argon} leads at $\bar S=65.2$ (95\% CI 56.3--73.9), while Claude Opus 5.5~\cite{anthropic2026opus55} and GPT-6.1 Sol~\cite{openai2026sol} tie at 60.2 (both 51.3--69.2).
These intervals overlap substantially, so task-bootstrap uncertainty does not resolve the ordering within the top three; the \emph{frontier cluster} is descriptive rather than a pairwise significance claim.
The next observed gap is nevertheless 22.4 points: Qwen3.8-Max-0902~\cite{qwen2026max0902} scores 37.8, then GLM-5.3-Flash~\cite{zai2026glm53flash}, Kimi K3~\cite{kimi2026k3}, MiMo-V2.6-Pro~\cite{mimo2026v26pro}, and Grok 4.7~\cite{xai2026grok47} sit between 24.0 and 28.9, while Seed 2.1 Pro~\cite{bytedance2026seed21} (12.6) and DeepSeek V4 Flash~\cite{deepseek2026v4} (12.1) stay near the floor.
Every model beats the best content-free lazy output (0.6/100) on 35--54 tasks, and the top three beat the direct classical operator (16.6/100) on 28--31 of its 34 applicable tasks; this is nontrivial but does not explain the gap to 100.

\paragraph{No Universal Model.}
\Cref{fig:teaser}-B and \cref{tab:main} show the family pattern.
Grouping (\famid{GRP}, mean 45.3) and motion (\famid{MOT}, 40.5) are the easiest families on average, while surfaces and light (\famid{SUR}, 28.0) and science and industry (\famid{SCI}, 28.6) are the hardest.
Strengths do not transfer: the mean pairwise Kendall rank correlation $\tau_{\mathrm{K}}$~\cite{kendall1938rank} between model family rankings is only 0.30.
Gemini 4 Argon leads \famid{GRP} (80.5), \famid{LAY} (94.7), and \famid{SCI} (62.1); GPT-6.1 Sol leads \famid{EDG} (76.1), \famid{SUR} (62.4), \famid{HID} (48.7), and \famid{REC} (73.6); Claude Opus 5.5 leads \famid{MOT} (64.6).
Low rank agreement and three family leaders show that the overall mean aggregates specializations, not one common ordering.
The same split appears inside one family. On \famid{LAY}, Gemini and Claude exceed 93, while MiMo-V2.6-Pro and Seed 2.1 Pro~\cite{bytedance2026seed21} score below 10.

\paragraph{Rendered Maps.}
\Cref{fig:qualitative} compares the four leading models on three real images at native aspect ratio.
On gland instances (\taskid{SCI02}), Gemini (65.0), GPT-6.1 Sol (59.0), and Claude (39.7) trace all 11 glands (foreground IoU 0.81--0.83), whereas Qwen draws 12 straight strips (IoU 0.51) and scores 3.5.
On document layout (\taskid{LAY01}), Gemini (99.3) and Claude (89.1) separate all 28 blocks, whereas GPT-6.1 Sol (18.0) and Qwen3.8-Max~\cite{qwen2026max0902} (25.1) label headers as body text or cross the gutter.
On the PCB (\taskid{SCI10}), Gemini (97.8) and Qwen (75.5) mark both the solder bridge and the thin trace; Claude (53.7) and GPT-6.1 Sol (54.2) mark only the bridge.
The coexistence of correct elements and local errors separates content recognition from geometric completeness.

\paragraph{Recognition before Geometry.}
\Cref{fig:results} separates the 242 checkpoints into coarse (21), mid-level (103), and pixel (118) tiers.
Within the same tasks, every model scores higher on coarse criteria than on pixel criteria: $\Delta_{\mathrm{crs},\mathrm{pix}}$ runs from $+14.0$ (GPT-6.1 Sol) to $+36.4$ (GLM-5.3-Flash), and $\Delta_{\mathrm{mid},\mathrm{pix}}$ from $+3.8$ to $+16.9$ (\cref{eq:tier-contrast}).
Within-task pairing rules out task mix, and the direct operator's $-2.1$ shows that a positive gap is not automatic; the pattern is consistent with loss from recognized structure to rendering, but cannot locate it among perception, memory, and program synthesis.
The top three still reach only 60.4--67.4\% credit on pixel checkpoints, against 72.1--84.7\% on coarse ones.
Real photographs widen the same difficulty: all ten models score lower on the 14 real-image tasks than on the 40 procedural ones (Gemini 40.8 vs.\ 73.8; Claude 25.4 vs.\ 72.4), while classical operators score higher on real photographs (24.3 vs.\ 13.5).
The reversal means source is not a universal difficulty order; because task composition also changes, it is descriptive rather than causal.

\paragraph{Comparison with Task-Specific Vision Models.}
With direct pixel access on the first released input per task, fixed classical pipelines cover 34 tasks (mean 16.2), while 34 task-specific methods comprise 7 exact, 12 proxy, and 15 heuristic adapters; their union gives all 54 tasks a fair direct-pixel comparator (details in the supplement).
Alignment reverses rankings: a retinal U-Net reaches 97.5 versus the best agent's 18.8, RightWayUp~\cite{ortusai2026rightwayup} 92.8 versus 84.7, and RAFT~\cite{teed2020raft} 60.3 versus 57.4.
LaMa~\cite{suvorov2022lama} reaches 93.6 on masked background completion and Swin2SR~\cite{liang2022swin2sr} 85.3 on text restoration, whereas generic 2D-to-3D transfer scores only 0.03--7.1 on hidden floor plans, novel views, and plane instances.
With a different method per task, the best-per-task mean of 24.6 is an oracle diagnostic, not one deployable system.

\subsection{In-Depth Analysis}
\label{sec:diag}
\label{sec:qual}
\label{sec:sensitivity}

\begin{figure}[htbp]
  \centering
  \includegraphics[width=0.68\linewidth]{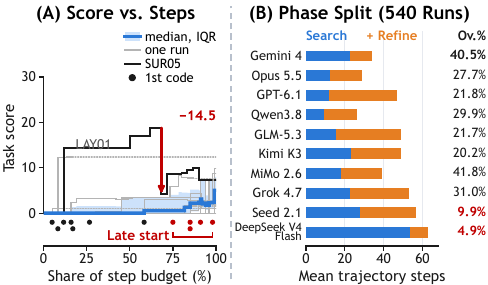}
  \caption{\textbf{Long-horizon trajectory dynamics.}
  (A)~12-task pilot (DeepSeek V4 Flash, thinking off): image-free re-executed score $S_\tau(t)$ over step budget $t/T_{\max}$ (blue: median/IQR; dots: first program; red: late starts at $\ge 75\%$ budget; black: \taskid{SUR05}, whose largest single edit loses 14.5 points).
  (B)~Mean trajectory phase split across all 540 main runs: visual search before first code ($0{\to}T_{\mathrm{1st}}$, blue) \vs closed-loop refinement ($T_{\mathrm{1st}}{\to}T$, orange) and self-overlay view share (Ov.\%).}
  \label{fig:pilot}
\end{figure}

\input{tables/diagnostics}

\paragraph{Trajectory Statistics.}
Across 540 main runs (22,680 steps, 28,131 views; \cref{fig:pilot}-B, \cref{tab:diagnostics}), the four leading models follow three stages: multi-scale search (crops in $68.4$--$87.8\%$ and grids in $64.7$--$80.1\%$ of views), first code at mean step $9.5$--$22.6$, and $4.6$--$7.6$ refinement iterations with $21.8$--$40.5\%$ self-overlay views.
MiMo never requests a grid, whereas DeepSeek writes first code at step $53.4$ and submits only $25.9\%$ of tasks.
These associations are not causal; Rep.\ in \cref{tab:diagnostics} measures byte-identical image-free re-execution.

\paragraph{Compute Budget.}
The budget diagnostic keeps the reasoning setting and the step limits fixed while doubling time/tokens for 13 timeout pairs, changing suite means by $+0.3$ to $+1.0$ (details in the supplement).
The response is not monotone (Doubao \taskid{SUR02} drops 16.0 points), so standard-budget scores remain primary.

\paragraph{Trajectory Case Studies.}
A 12-task DeepSeek pilot re-executes 78 snapshots without images (\cref{fig:pilot}-A): $57.1\%$ of steps precede first code, and step-20 snapshots reach $39.9\%$ of final scores ($\mathrm{AUC}_\tau/S_\tau=0.56$).
Together with \cref{fig:qualitative}, three regimes emerge:
\begin{itemize}
\item \textbf{Closed-loop grounding (\taskid{SCI10})}: Gemini uses 70 views and four overlays to reach 97.8; Claude uses 7 views and misses the thin trace (53.7).
\item \textbf{Non-monotonic editing (\taskid{SUR05}, \taskid{MOT01})}: five pilot edits improve and four degrade; final scores fall from peaks by 14.5 and 3.3 points.
\item \textbf{Contour and layout geometry (\taskid{SCI02}, \taskid{LAY01})}: Qwen replaces gland contours with straight strips (3.5), and column alignment separates Gemini/Claude (99.3/89.1) from GPT/Qwen (18.0/25.1).
\end{itemize}

%% file: tables/mainresults.tex
\begin{table}[htbp]
\centering
\footnotesize
\setlength{\tabcolsep}{2.0pt}
\renewcommand{\arraystretch}{1.12}
\newcolumntype{Y}{>{\centering\arraybackslash}X}
\caption{\textbf{Main results on \bench} (0--100; image-free re-execution across all 54 tasks).
Overall weights all 54 tasks equally; the 95\% CI is a percentile bootstrap over tasks~\cite{efron1994bootstrap}.
C$-$P is the within-task coarse-minus-pixel checkpoint gap (\cref{eq:tier-contrast}).
Think is the requested reasoning effort: high for eight models, adaptive for Claude (capped at high), and on for MiMo, whose endpoint exposes only enabled/disabled.
Best and second-best family and overall scores are shown in \best{red} and \second{blue}.}
\label{tab:main}
\begin{tabularx}{\linewidth}{@{}l c *{8}{Y} c c c@{}}
\toprule
\multirow{2.3}{*}{\textbf{Model}} & \multirow{2.3}{*}{\textbf{Think}} & \multicolumn{8}{c}{\textbf{Task Family (54 Tasks across 8 Families)}} & \multirow{2.3}{*}{\textbf{Overall}} & \multirow{2.3}{*}{\textbf{95\% CI}} & \multirow{2.3}{*}{\textbf{C$-$P}} \\
\cmidrule(lr){3-10}
 & & \textbf{\famid{EDG}} & \textbf{\famid{GRP}} & \textbf{\famid{SUR}} & \textbf{\famid{HID}} & \textbf{\famid{MOT}} & \textbf{\famid{REC}} & \textbf{\famid{LAY}} & \textbf{\famid{SCI}} & & & \\[-0.25ex]
 & & {\scriptsize (6)} & {\scriptsize (7)} & {\scriptsize (6)} & {\scriptsize (7)} & {\scriptsize (8)} & {\scriptsize (7)} & {\scriptsize (3)} & {\scriptsize (10)} & & & \\
\midrule
Gemini 4 Argon~\cite{google2026gemini4argon} & high & \second{75.7} & \best{80.5} & \second{58.4} & 46.5 & \second{62.5} & \second{60.5} & \best{94.7} & \best{62.1} & \best{65.2} & [56.3, 73.9] & $+24.6$ \\
Claude Opus 5.5~\cite{anthropic2026opus55} & adaptive & 65.5 & \second{73.5} & 53.9 & \second{47.2} & \best{64.6} & 56.9 & \second{93.2} & \second{49.7} & \second{60.2} & [51.3, 69.2] & $+19.2$ \\
GPT-6.1 Sol~\cite{openai2026sol} & high & \best{76.1} & 60.9 & \best{62.4} & \best{48.7} & 60.3 & \best{73.6} & 68.7 & 45.1 & \second{60.2} & [51.3, 69.2] & $+14.0$ \\
Qwen3.8-Max-0902~\cite{qwen2026max0902} & high & 43.9 & 47.4 & 28.5 & 34.8 & 44.4 & 35.4 & 39.2 & 31.1 & 37.8 & [30.7, 45.2] & $+35.8$ \\
GLM-5.3-Flash~\cite{zai2026glm53flash} & high & 23.4 & 44.2 & 17.5 & 28.1 & 32.4 & 35.6 & 28.2 & 21.5 & 28.9 & [22.2, 36.0] & $+36.4$ \\
Kimi K3~\cite{kimi2026k3} & high & 21.3 & 37.2 & 21.0 & 26.5 & 28.5 & 33.2 & 18.1 & 24.7 & 27.1 & [20.8, 33.8] & $+25.5$ \\
MiMo-V2.6-Pro~\cite{mimo2026v26pro} & on & 1.5 & 48.7 & 6.8 & 25.5 & 50.9 & 34.8 & 9.3 & 21.1 & 27.0 & [18.8, 35.8] & $+14.9$ \\
Grok 4.7~\cite{xai2026grok47} & high & 27.3 & 31.1 & 14.6 & 21.6 & 36.2 & 18.4 & 26.7 & 17.5 & 24.0 & [18.3, 29.9] & $+35.7$ \\
Doubao Seed 2.1 Pro~\cite{bytedance2026seed21} & high & 23.9 & 14.3 & 8.1 & 11.1 & 13.7 & 18.1 & 7.3 & 5.4 & 12.6 & [9.0, 16.9] & $+26.6$ \\
DeepSeek V4 Flash~\cite{deepseek2026v4} & high & 12.4 & 15.1 & 8.5 & 18.9 & 11.6 & 13.5 & 8.0 & 7.9 & 12.1 & [8.0, 16.7] & $+16.8$ \\
\bottomrule
\end{tabularx}
\end{table}

%% file: tables/diagnostics.tex
\begin{table}[htbp]
\centering
\small
\setlength{\tabcolsep}{4.5pt}
\renewcommand{\arraystretch}{1.10}
\caption{\textbf{Long-horizon trajectory, visual tool-use, program synthesis, and outcome statistics across all 54 tasks (540 runs).}
Steps/1st/Rev.: mean total steps, step of first program synthesis, and code write/run iterations; Views/Crop/Grid/Ov.: mean views per task, zoomed-crop share (\%), coordinate-grid share (\%), and closed-loop self-overlay share (\%); Lits.: mean numeric literals per program; Sub./Rep.: explicit submission and stateless image-free reproduction rates (\%).}
\label{tab:diagnostics}
\begin{tabularx}{\linewidth}{@{}l @{\extracolsep{\fill}} rrr rrrr r rr@{}}
\toprule
\multirow{2.2}{*}{\textbf{Model}} & \multicolumn{3}{c}{\textbf{Trajectory}} & \multicolumn{4}{c}{\textbf{Visual tool loop}} & \textbf{Code} & \multicolumn{2}{c}{\textbf{Outcome}} \\
\cmidrule(lr){2-4} \cmidrule(lr){5-8} \cmidrule(lr){9-9} \cmidrule(l){10-11}
 & Steps & 1st & Rev. & Views & Crop & Grid & Ov. & Lits. & Sub. & Rep. \\
\midrule
Gemini 4 Argon~\cite{google2026gemini4argon}      & 34.1 & 22.6 & 7.6 & 47.6 & 74.8 & 80.1 & 40.5 & 486 &  90.7 &  94.4 \\
Claude Opus 5.5~\cite{anthropic2026opus55}     & 29.0 & 12.4 & 4.6 & 25.6 & 77.3 & 73.6 & 27.7 & 259 & 100.0 &  98.1 \\
GPT-6.1 Sol~\cite{openai2026sol}         & 46.9 & 11.9 & 6.5 & 94.1 & 87.8 & 79.4 & 21.8 & 496 &  98.1 &  98.1 \\
Qwen3.8-Max-0902~\cite{qwen2026max0902}    & 26.2 &  9.5 & 7.6 & 23.3 & 68.4 & 64.7 & 29.9 & 671 & 100.0 & 100.0 \\
GLM-5.3-Flash~\cite{zai2026glm53flash}       & 49.1 & 15.5 & 9.4 & 52.9 & 81.7 & 60.0 & 21.7 & 322 &  79.6 &  96.3 \\
Kimi K3~\cite{kimi2026k3}             & 48.8 & 23.3 & 6.9 & 48.6 & 76.3 & 62.4 & 20.2 & 259 &  96.3 &  92.6 \\
MiMo-V2.6-Pro~\cite{mimo2026v26pro}       & 39.1 & 18.1 & 5.1 & 27.2 & 71.6 &  0.0 & 41.8 & 217 &  64.8 &  64.8 \\
Grok 4.7~\cite{xai2026grok47}            & 53.2 & 22.7 & 9.0 & 86.6 & 86.5 & 71.8 & 31.0 & 388 &  87.0 &  96.3 \\
Doubao Seed 2.1 Pro~\cite{bytedance2026seed21}  & 56.9 & 27.9 & 6.8 & 52.6 & 76.8 & 75.8 &  9.9 & 195 &  68.5 &  92.6 \\
DeepSeek V4 Flash~\cite{deepseek2026v4}   & 63.1 & 53.4 & 3.2 & 99.5 & 83.8 & 87.1 &  4.9 & 209 &  25.9 &  87.0 \\
\bottomrule
\end{tabularx}
\end{table}

%% file: sections/05-conclusion.tex
\section{Conclusion}
\label{sec:conclusion}

\bench evaluates generalist multimodal agents on classic vision tasks under one rule: input pixels reach only the agent's vision, and the scored map is drawn by a program re-executed without images.
\emph{Even the strongest agents remain far from solving the suite, and every model earns more credit for recognizing scene content than for delineating it at pixel accuracy.}
No model leads every family; successful runs rely on iterative search and closed-loop verification, although late edits can still hurt.
\emph{Limitations.} Most tasks hold one or two inputs and each model--task pair runs once, so intervals reflect task selection rather than run-to-run variance; prompts are English only, real-image tasks may overlap pretraining data, and fixed thresholds encode one annotator-defined standard.
The closed-book protocol isolates perception, not tool-augmented deployment.